\documentclass{article}

     \usepackage[final]{neurips_2018}

\usepackage[utf8]{inputenc} % allow utf-8 input
\usepackage[T1]{fontenc}    % use 8-bit T1 fonts
\usepackage{hyperref}       % hyperlinks
\usepackage{url}            % simple URL typesetting
\usepackage{booktabs}       % professional-quality tables
\usepackage{amsfonts}       % blackboard math symbols
\usepackage{nicefrac}       % compact symbols for 1/2, etc.
\usepackage{microtype}      % microtypography
\usepackage{amsmath,amsthm,amssymb}
\usepackage{graphicx}
\usepackage{wrapfig}
\usepackage{booktabs}

\title{Hybrid Macro/Micro Level Backpropagation for Training Deep Spiking Neural Networks}

\author{
  Yingyezhe Jin\\
  Texas A\&M University\\
  College Station, TX 77843 \\
  \texttt{jyyz@tamu.edu} \\
  \And
  Wenrui Zhang\\
  Texas A\&M University\\
  College Station, TX 77843 \\
  \texttt{zhangwenrui@tamu.edu} \\
  \And
  Peng Li\\
  Texas A\&M University\\
  College Station, TX 77843 \\
  \texttt{pli@tamu.edu} \\
}

\begin{document}

\maketitle
\vspace{-4mm}
\begin{abstract}
\vspace{-3mm}
 Spiking neural networks (SNNs) are positioned to enable spatio-temporal information processing and ultra-low power event-driven neuromorphic hardware. However,  SNNs are yet to reach the same performances of conventional deep artificial neural networks (ANNs), a long-standing challenge due to complex dynamics and non-differentiable spike events encountered in training. The existing SNN error backpropagation (BP) methods are limited in terms of scalability, lack of proper handling of spiking discontinuities, and/or mismatch between the rate-coded loss function and computed gradient. We present a hybrid macro/micro level backpropagation  (HM2-BP) algorithm for training multi-layer SNNs. The temporal effects are precisely captured by the proposed \emph{\underline{s}pike-train level \underline{p}ost-\underline{s}ynaptic \underline{p}otential} (S-PSP) at the microscopic level.  The rate-coded errors are defined at the macroscopic level,  computed and back-propagated across both macroscopic and microscopic levels.  Different from existing BP methods, HM2-BP directly computes the gradient of the rate-coded loss function w.r.t tunable parameters. We evaluate the proposed HM2-BP algorithm by training deep fully connected and convolutional SNNs based on the static MNIST~\cite{lecun1998gradient} and dynamic neuromorphic N-MNIST~\cite{orchard2015converting}. HM2-BP achieves an accuracy level of $99.49\%$ and $98.88\%$ for MNIST and N-MNIST, respectively, outperforming the best reported performances obtained from the existing SNN BP algorithms. Furthermore, the HM2-BP produces the highest accuracies based on SNNs for the EMNIST~\cite{cohen2017emnist} dataset, and leads to high recognition accuracy for the 16-speaker spoken English letters of TI46 Corpus~\cite{TI46}, a challenging spatio-temporal speech recognition benchmark for which no prior success based on SNNs was reported. It also achieves competitive performances surpassing those of conventional deep learning models when dealing with asynchronous spiking streams.% Our source code is available online\footnote{https://github.com/jinyyy666/mm-bp-snn}.
\end{abstract}
\vspace{-4mm}
\section{Introduction}
\vspace{-3mm}
In spite of recent success in deep neural networks (DNNs)~\cite{collobert2008unified, hinton2012deep, krizhevsky2012imagenet}, it is believed that biological brains operate rather differently. Compared with DNNs that lack processing of spike timing  and event-driven operations, biologically realistic spiking neural networks (SNNs)~\cite{izhikevich2008large, maass1997networks} provide a promising paradigm for exploiting spatio-temporal patterns for added computing power, and enable ultra-low power event-driven neuromorphic hardware~\cite{benjamin2014neurogrid, esser2015backpropagation, merolla2014million}. There are theoretical evidences supporting that SNNs possess greater computational power  over traditional artificial neural networks (ANNs)~\cite{maass1997networks}. SNNs are yet to achieve a performance level on a par with deep ANNs for practical applications. The error backpropagation~\cite{rumelhart1986learning} is very successful in training ANNs. Attaining the same success of backpropagation (BP) for SNNs is challenged by two fundamental issues: complex temporal dynamics and non-differentiability of discrete spike events. 
%due to two fundamental difficulties of dealing with complex temporal effects and spike discontinuity. The error-backpropagation~\cite{rumelhart1986learning}, which has been successful in training ANNs, can be a good candidate for training SNNs. However, the complicated temporal dynamics and non-differentiable discrete spike events are not congruent with the differentiability and continuity prerequisites of conventional backpropagation, making it very challenging to train SNNs. 
%closely resemble the brain behavior. In contrast of DNNs that lack timing information, SNNs exploit spatio-temporal patterns in spike timing for added computing power. Event-driven processing characteristics of SNNs are favorable over DNNs as it facilitates energy-efficient neuromorphic hardware platforms~\cite{benjamin2014neurogrid, esser2015backpropagation, merolla2014million}. 

\textbf{Problem Formulation:} As a common practice in SNNs, the rate coding is often adopted to define a loss for each training example at the output layer \cite{lee2016training, wu2017spatio}
\begin{equation}
    \label{rate_obj}
    E = \frac{1}{2} ||\mathbf{o} - \mathbf{y}||^2_2,
\end{equation}
where $\mathbf{o}$ and $\mathbf{y}$  are vectors  specifying the actual and desired (label) firing counts  of the output neurons. Firing counts are determined by the underlying firing events, which are adjusted discretely by tunable weights, resulting in great challenges in computing the gradient of the loss with respect to the weights.  

\textbf{Prior Works:} There exist approaches that stay away from the SNN training challenges by first training an ANN and then approximately converting it to an SNN ~\cite{diehl2015fast, esser2015backpropagation, hunsberger2015spiking, o2013real}. \cite{o2016deep} takes a similar approach which treats spiking neurons almost like non-spiking ReLU units. The accuracy of those methods may be severely compromised because of imprecise representation of timing statistics of spike trains. Although the latest ANN-to-SNN conversion approach~\cite{rueckauer2017conversion} shows promise, the problem of direct training of SNNs remains unsolved.

%One of the earliest attempts to bridge the gap between discontinuity of SNNs and backpropagation is the SpikeProp algorithm~\cite{bohte2002error}, which is specific for temporal learning and derived on the simple assumption that the spike response at output spike instant is linear on time. However, the SpikeProp is restricted to single-spike learning, resulting in limited learning efficiency, and has not yet been successful in solving any real-world benchmarks. Some other approaches circumvent this difficulty by indirectly building a continuous and differentiable substitute, such as a conventional ANN or a probabilistic model, training the obtained structure via backpropagation and then converting it into a SNN with the same topology and weights~\cite{diehl2015fast, esser2015backpropagation, hunsberger2015spiking, o2013real}. Unfortunately, the accuracy of those methods is severely compromised because of imprecise representation and failure of capturing spike stream statistics. 

The SpikeProp algorithm ~\cite{bohte2002error} is the first attempt to train an SNN by operating on discontinuous spike activities. It specifically targets temporal learning for which derivatives of the loss w.r.t. weights are explicitly derived. However, SpikeProp is very much limited to single-spike learning, and its successful applications to realistic benchmarks have not been demonstrated. Similarly, \cite{zenke2018superspike} proposed a temporal training rule for understanding learning in SNNs. More recently, the backpropagation approaches of ~\cite{lee2016training} and~\cite{wu2017spatio} have shown competitive performances. Nevertheless,  \cite{lee2016training} lacks explicit consideration of temporal correlations of neural activities. Furthermore, it does not handle discontinuities occurring at spiking moments by treating them as noise while only computing the error gradient for the remaining smoothed membrane voltage waveforms instead of the rate-coded loss. \cite{wu2017spatio} addresses the first limitation of~\cite{lee2016training} by performing  BPTT~\cite{werbos1990backpropagation} to capture temporal effects.  However, similar to~\cite{lee2016training}, the error gradient is computed for the continuous membrane voltage waveforms resulted from smoothing out all spikes, leading to inconsistency w.r.t the rate-coded loss function. In summary, the existing SNNs BP algorithms have three major limitations: i) suffering from limited learning scalability~\cite{bohte2002error},  ii) either staying away from spiking discontinuities (e.g. by treating spiking moments as noise~\cite{lee2016training}) or deriving the error gradient based on the smoothed membrane waveforms~\cite{lee2016training, wu2017spatio}, and therefore iii) creating a mismatch between the computed gradient and targeted rate-coded loss~\cite{lee2016training,wu2017spatio}.

\textbf{Paper Contributions:} We derive the gradient of the rate-coded error defined in (\ref{rate_obj})  by decomposing each derivative into two components
\begin{equation}
    \begin{aligned}
    \frac{\partial E}{\partial w_{ij}} = \underbrace{\frac{\partial E}{\partial a_i}}_\text{bp over firing rates} \times \underbrace{\frac{\partial a_i}{\partial w_{ij}}}_\text{bp over spike trains},
    \end{aligned}
\end{equation}
\begin{wrapfigure}{r}{0.3\textwidth}
    \vspace{-6.5mm}
    \begin{center}
        \includegraphics[width=0.3\textwidth]{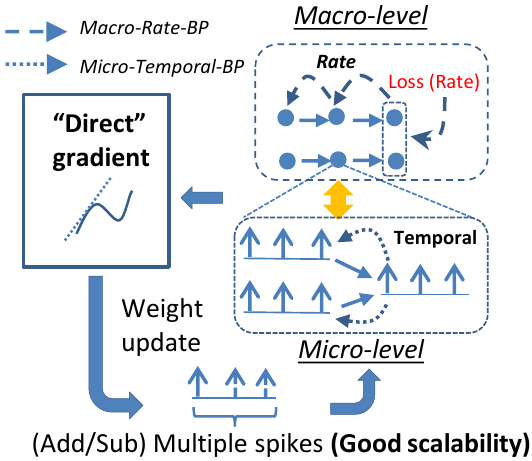}
    \end{center}
    \vspace{-4mm}
    \caption{Hybrid macro-micro level backpropagation.}
    \label{high_level}
    \vspace{-2mm}
\end{wrapfigure}
where $a_i$ is the (weighted) aggregated membrane potential for the post-synaptic neuron $i$ per (\ref{ai_formula}). As such, we propose a novel hybrid macro-micro level backpropagation (HM2-BP) algorithm which performs error backpropagation across two levels: 1) backpropagation over firing rates (\textbf{\emph{macro-level}}), 2) backpropagation over spike trains (\textbf{\emph{micro-level}}), and 3) backpropagation based on interactions between the two levels, as illustrated in Fig.~\ref{high_level}.

At the microscopic level, for each pre/post-synaptic spike train pair, we precisely compute the \emph{\underline{s}pike-train level \underline{p}ost-\underline{s}ynaptic \underline{p}otential}, referred to as \emph{S-PSP} throughout this paper, to account for the temporal contribution of the given pre-synaptic spike train to the firings of the post-synaptic neuron based on  exact spike times. 
At the macroscopic level, we back-propagate the errors of the defined rate-based loss by aggregating the effects of spike trains  on each neuron's firing count via the use of S-PSPs, and leverage this as a practical way of linking  spiking events to firing rates.  To assist backpropagation, we further propose a decoupled model of the S-PSP for disentangling the effects of firing rates and spike-train timings to allow differentiation of the S-PSP  w.r.t. pre and post-synaptic firing rates at the micro-level. As a result, our HM2-BP approach is able to evaluate the direct impact of weight changes on the rate-coded loss function. Moreover, the resulting weight updates in each training iteration can lead to introduction or disappearance of multiple spikes. 

We evaluate the proposed BP algorithm by training deep fully connected and convolutional SNNs based on the static MNIST~\cite{lecun1998gradient}, dynamic neuromorphic N-MNIST~\cite{orchard2015converting}, and EMNIST~\cite{cohen2017emnist} datasets. Our BP algorithm achieves an accuracy level of $99.49\%$, $98.88\%$ and $85.57\%$ for MNIST, N-MNIST and EMNIST, respectively, outperforming the best reported performances obtained from the existing SNN BP algorithms. Furthermore, our algorithm achieves high recognition accuracy of 90.98\% for the 16-speaker spoken English letters of TI46 Corpus~\cite{TI46}, a challenging spatio-temporal speech recognition benchmark for which no prior success based on SNNs was reported.

\section{Hybrid Macro-Micro Backpropagation}
\label{main}
The complex dynamics generated by spiking neurons and non-differentiable spike impulses are two fundamental bottlenecks for training SNNs using backpropagation. We address these difficulties at both macro and micro levels. 

%this work captures temporal effects at the micro-level by using S-PSPs and deals with spiking discontinuities at macro/micro levels, allowing backpropagation of the error of the rate-coded objective function through computations taking place at both macro and micro levels. A decoupled model for the S-PSP is proposed to enable the error backpropagation at the micro-level. 

%we use \emph{total post-synaptic potential effect} (S-PSP), which is continuous in nature, as a underlying differentiable variable to establish the error backward pass on the spike train level, avoiding the above resulted undesirable approximations for handling the discontinuity inborn in the spiking neuron model. By means of S-PSP, we can also quantify the effect of each tunable weight to a neuron final firing rate, which has been somehow neglected in the previous works.  %(e.g., fixed-amount-reset and treating discontinuity at spike times as noise for~\cite{lee2016training}; an ``iterative'' LIF model and approximation of spiking event derivative for~\cite{wu2017spatio}) The previous works~\cite{lee2016training, wu2017spatio} mainly focus on approximations on single spike characteristics and neuron models to make SNNs eligible for error-backpropagation.

\subsection{Micro-level Computation of Spiking Temporal Effects}
%We begin with investigating the individual build block (i.e., neuron model) of SNNs and how the temporal information shall be grasped from the micro-level perspective. 

The leaky integrate-and-fire (LIF) model is one of the most prevalent choices for describing dynamics of spiking neurons, where the neuronal membrane voltage $u_i(t)$ at time $t$ for the neuron $i$ is given by
\begin{equation}
    \label{lif}
    \tau_m \frac{u_i(t)}{dt} = -u_i(t) + R~I_i(t),
\end{equation}
where $I_i(t)$ is the input current, $R$  the effective leaky resistance, $C$  the effective membrane capacitance, and $\tau_m = RC$  the membrane time constant.  A spike is generated when  $u_i(t)$ reaches the threshold $\nu$.  After that $u_i(t)$ is reset to the resting potential $u_r$, which equals to $0$ in this paper. Each post-synaptic neuron $i$ is driven by a post-synaptic current of the following general form 
\begin{equation}
    \label{It}
    I_i(t) = \sum_j w_{ij} \sum_f \alpha (t - t_j^{(f)}),
\end{equation}
where $w_{ij}$ is the weight of the synapse from the pre-synaptic neuron $j$ to the neuron $i$, $t_j^{(f)}$ denotes a particular firing time of the neuron $j$. We adopt a first order synaptic model with time constant $\tau_s$
\begin{equation}
    \alpha(t) = \frac{q}{\tau_s} \exp\left(-\frac{t}{\tau_s} \right) H(t),
\end{equation}
where $H(t)$ is the Heaviside step function, and $q$  the total charge injected into the post-synaptic neuron $i$ through a synapse of a weight of 1.  Let $\hat{t}_i$ denote the last firing time of the neuron $i$ w.r.t time $t$: $\hat{t}_i = \hat{t}_i(t) = \max\{t_i | t_i^{(f)} < t \}$. Plugging~(\ref{It}) into~(\ref{lif}) and integrating~(\ref{lif}) with $u(\hat{t}_i) = 0$ as its initial condition, we  map the LIF model to the Spike Response Model (SRM)~\cite{gerstner2002spiking}
\begin{equation}
    \label{srm}
        %u_i(t) = & ~ u_r \exp(-\frac{t - \hat{t}_i}{\tau_m}) + \sum_j w_{ij} \sum_f \epsilon(t - \hat{t}_i, t - t_j^{(f)}),
    u_i(t) = \sum_j w_{ij} \sum_f \epsilon\left(t - \hat{t}_i, t - t_j^{(f)}\right),
\end{equation}
with
\begin{equation}
    \label{psp_resp}
    \epsilon(s, t) = ~ \frac{1}{C} \int_0^{s} \exp\left(-\frac{t^{\prime}}{\tau_m}\right)~\alpha \left(t - t^{\prime}\right)~\mathrm{d}t^{\prime}.
\end{equation}
Since  $q$ and $C$ can be absorbed into the synaptic weights, we set $q = C = 1$.  Integrating~(\ref{psp_resp}) yields
\begin{equation}
    \label{psp}
    \epsilon(s, t) = \frac{\exp(-\max(t - s, 0) /\tau_s)}{1 - \frac{\tau_s}{\tau_m}} ~~ \left[\exp\left(-\frac{\min(s, t)}{\tau_m}\right) - \exp\left(-\frac{\min(s, t)}{\tau_s}\right)\right]H(s)H(t).
\end{equation}
\begin{wrapfigure}{r}{0.25\textwidth}
    \vspace{-4mm}
    \begin{center}
        \includegraphics[width=0.25\textwidth]{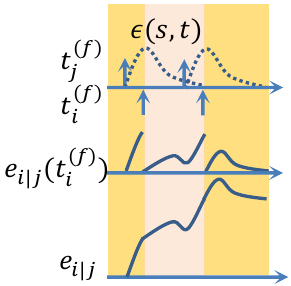}
    \end{center}
    \vspace{-4mm}
    \caption{The computation of the S-PSP.}
    \label{pair_demo}
\end{wrapfigure}
$\epsilon$ is interpreted as the normalized (by synaptic weight) \emph{post-synaptic potential}, which is evoked by a single firing spike of the pre-synaptic neuron $j$. 

For any time $t$, the exact "contribution" of  the neuron $j$'s spike train  to the neuron $i$'s post-synaptic potential  is given by summing~(\ref{psp}) over all pre-synaptic spike times $t_j^{(f)}$, $t_j^{(f)} < t$. We particularly concern the contribution right before each post-synaptic firing time $t_i^{(f)}$ when $u_i(t_i^{(f)}) = \nu$, which we denote by $e_{i | j}(t_i^{(f)})$. Summing $e_{i | j}(t_i^{(f)})$ over all post-synaptic firing times gives the \emph{total} contribution of the neuron $j$'s spike-train to the firing activities of the neuron $i$ as shown in Fig.~\ref{pair_demo}
\begin{equation}
    \label{effect}
    e_{i | j} = \sum_{t_i^{(f)}} \sum_{t_j^{(f)}} \epsilon(t_i^{(f)} - \hat{t}_i^{(f)}, t_i^{(f)} - t_j^{(f)}),
\end{equation}
where $\hat{t}_i^{(f)} =\hat{t}_i^{(f)} (t_i^{(f)})$ denotes the last post-synaptic firing time before $t_i^{(f)}$. 

Importantly, we refer to $e_{i | j}$ as the (normalized) \emph{spike-train level post-synaptic potential} (S-PSP).  As its name suggests, S-PSP characterizes  the aggregated influence of the pre-synaptic neuron on the post-synaptic neuron's firings at the level of spike trains, providing a basis for relating firing counts to spike events and enabling scalable SNN training that adjusts spike trains rather than single spikes.   Clearly, each S-PSP $e_{i|j}$ depends on both rate and temporal information of the pre/post spike trains. To assist the derivation of our BP algorithm, we make the dependency of $e_{i|j}$ on the pre/post-synaptic firing counts $o_i$ and $o_j$ explicit although $o_i$ and $o_j$ are already embedded in the spike trains
%Before deriving the proposed backpropagation rule, it is also worthwhile mentioning that the S-PSP $e_{j | i}$ can be expressed in terms of the rate information of both pre- and post-synaptic neuron:
\begin{equation}
    \label{effect_formula}
    e_{i | j} = f(o_j, o_i, \mathbf{{t}}_j^{(f)}, \mathbf{t}_i^{(f)}),
\end{equation}
where  $\mathbf{t}_j^{(f)}$ and $\mathbf{t}_i^{(f)}$ represent the pre and post-synaptic timings, respectively. %(\ref{effect_formula}) is very essential for computing the derivative of $a_i$ with respect to the weight $w_{ij}$ as we will show in Section~\ref{sec:macro}. 
%The S-PSP $e_{i | j}$ describes the total temporal effect brought by the pre-synaptic spike events $\sum t_j^{(f)}$ upon the membrane potential of the post-synaptic neuron $i$ given $\sum t_i^{(f)}$ as the resulted post-synaptic firing events.% Note that the S-PSP eliminates the necessity to individually handle the spike discontinuity. %Notice that the effect of refractory periods can be considered by slightly modifying the timings.  $=\hat{t}_i^{(f)} (t_i^{(f)})$
%When $u_i(t)$ reaches the threshold $\nu$, we can \emph{exactly} compute how much ``contribution'' the spike sequence of the neuron $j$ has upon the current firing, which is given by the summation of~(\ref{psp}) over pre-synaptic spike times $\sum t_j^{(f)}$. We then conduct this computation over all firing times $t_i^{(f)} : u_i(t_i^{(f)}) = \nu$ of the post-synaptic neuron $i$, and get the normalized \emph{spike-train level post-synaptic potential} (S-PSP) (see Fig.~\ref{pair_demo}) resulted from the pre-synaptic spike sequence as
Summing the weighted S-PSPs from all pre-synaptic neurons results in the \emph{total post-synaptic potential} (T-PSP) $a_i$, which is directly correlated to the neuron $i$'s firing count
\begin{equation}
    \label{ai_formula}
    a_i = \sum_j w_{ij}~ e_{i|j}.
\end{equation}

%Via the S-PSP $e_{j | i}$, we establish the precise relationship between the fire count $o_i$ and the synaptic weight $w_{ij}$ given the pre and post-synaptic spike sequence, which is the foundation for our error backpropagation rule. Notice that we do not need any approximation for the individual spike response as~\cite{wu2017spatio} did, because the information of the response is completely captured by the S-PSP, obtained through the spike train level computation as shown in~(\ref{psp}) given two spike trains.

\subsection{Error Backpropagation at Macro and Micro Levels}
\label{sec:macro}
%We now describe how our rate-coded error is defined at the macro-level and how this error backpropagation can be done in both macro and micro levels for each layer. 
%At the macro-level, we adopt the rate coding, and the break-down loss for each sample is defined as the mean square error
It is evident that the total post-synaptic potential $a_i$ must be no less than the threshold $\nu$ in order to make the neuron $i$ fire at least once, and the total firing count is $\left\lfloor\frac{a_i}{\nu}\right\rfloor$. 
We relate the firing count $o_i$ of the neuron $i$ to $a_i$ approximately by
\begin{equation}
    \label{tf_formula}
    o_i = g(a_i) = \left\lfloor\frac{a_i}{\nu}\right\rfloor = \left\lfloor\frac{\sum_j w_{ij}~ e_{i|j}}{\nu}\right\rfloor \approx \frac{\sum_j w_{ij}~ e_{i|j}}{\nu},
\end{equation}
where the rounding error would be insignificant when $\nu$ is small. Despite that (\ref{tf_formula}) is linear in S-PSPs,  it is the interaction between the S-PSPs through nonlinearities hidden in the micro-level LIF model that leads to a given firing count $o_i$.  Missing from the existing works~\cite{lee2016training,wu2017spatio}, (\ref{tf_formula}) serves as an important bridge connecting the aggregated micro-level temporal effects with the macro-level count of discrete firing events.   In a vague sense, $a_i$ and $o_i$ are analogies to pre-activation and activation in the traditional ANNs, respectively, although they are not directly comparable.  (\ref{tf_formula}) allows for rate-coded error backpropagation on top of discrete spikes across the macro and micro levels.

Using (\ref{tf_formula}), the macro-level rate-coded loss of (\ref{rate_obj}) is rewritten as
\begin{equation}\label{eqn_loss}
    E = \frac{1}{2} ||\mathbf{o} - \mathbf{y}||^2_2 = \frac{1}{2} ||g(\mathbf{a}) - \mathbf{y}||^2_2,
\end{equation}
where $\mathbf{y}$, $\mathbf{o}$ and $\mathbf{a}$ are vectors specifying the desired firing counts (label vector), the actual firing counts, and the weighted sums of S-PSP of the output neurons, respectively. 
%By adopting (\ref{tf_formula}), we link the discontinuous spiking events to firing counts in an aggregated manner to deal with the individual spike discontinuity and obtain rate-based error gradients. %and each element $y_i$ is a binary class label that is the desirable firing level $d_{lvl}$ if the class is present and the undesirable level $\Bar{d}_{lvl}$ otherwise. By adopting (\ref{tf_formula}), we manage spike discontinuity to obtain rate-based error gradients.% According to (\ref{tf_formula}), we know that this rate-based error $E$ is a function of synaptic weights $w_{ij}$. 
We now derive the gradient of $E$ w.r.t  $w_{ij}$ at each layer of an SNN.

\textbullet\ \textbf{Output layer:} For the $i_{th}$ neuron in the output layer $m$, we have
\begin{wrapfigure}{r}{0.28\textwidth}
    \begin{center}
        \includegraphics[width=0.28\textwidth]{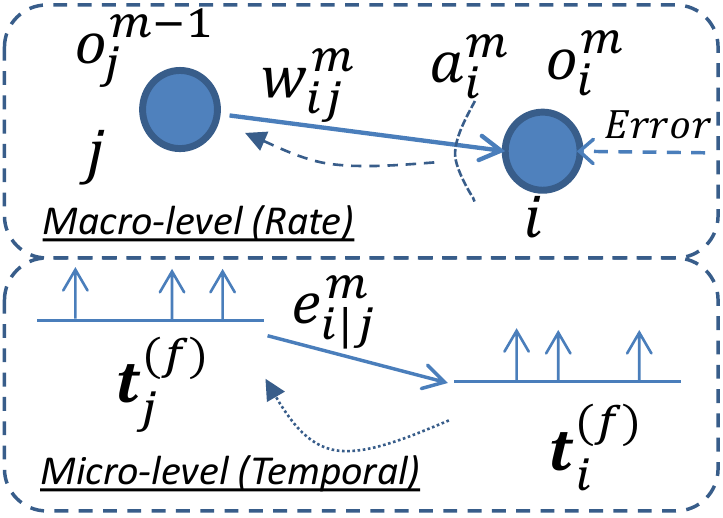}
    \end{center}
    \vspace{-5mm}
    \caption{Macro/micro backpropagation in the output layer.}
    \label{output_bp_img}
\end{wrapfigure}
\begin{equation}
    \label{dEdW}
    \begin{aligned}
    \frac{\partial E}{\partial w_{ij}} = \underbrace{\frac{\partial E}{\partial a_i^m}}_\text{macro-level bp} \times \underbrace{\frac{\partial a^m_i}{\partial w_{ij}}}_\text{micro-level bp},
    \end{aligned}
\end{equation}
where variables associated with neurons in the layer $m$ have $m$ as the superscript. As shown in Fig.~\ref{output_bp_img}, the first term of (\ref{dEdW}) represents the macro-level backpropagation of the rate-coded error with the second term being the micro-level error backpropagation. From (\ref{eqn_loss}), the macro-level error backpropagation is given by
\begin{equation}
    \label{delta_output}
    \delta^m_i = \frac{\partial E}{\partial a_i^m} =(o^m_i - y^m_i)~g^\prime(a^m_i) = \frac{o^m_i - y^m_i}{\nu}.
\end{equation}
Similar to the conventional backpropagation, we use $\delta^m_i$ to denote the back propagated error. According to (\ref{ai_formula}) and (\ref{effect_formula}), $a^m_i$ can be unwrapped as
\begin{equation}
    \label{ai_expand}
    a^m_i = \sum_{j = 1}^{r^{m-1}} w_{ij} ~ e^m_{i | j} = \sum_{j = 1}^{r^{m-1}} w_{ij} ~ f(o_j^{m-1}, o_i^{m}, \mathbf{t}_j^{(f)}, \mathbf{t}_i^{(f)}),
\end{equation}
where $r^{m-1}$ is the number of neurons in the $(m-1)_{th}$ layer.
Differentiating (\ref{ai_expand}) and making use of (\ref{tf_formula}) leads to the micro-level error propagation based on the total post-synaptic potential (T-PSP) $a^m_i$
\begin{equation}
    \label{dadw}
    \frac{\partial a^m_i}{\partial w_{ij}} = \frac{\partial }{\partial w_{ij}}\left(\sum_{j = 1}^{r^{m-1}} w_{ij} ~ e^m_{i | j}\right) = e^m_{i | j} + \sum_{l=1}^{r^{m-1}} w_{il}\frac{\partial e^m_{i | l}}{\partial o^m_i}~\frac{\partial o^m_i}{\partial w_{ij}} = e^m_{i | j} + \frac{e^m_{i | j}}{\nu}~\sum_{l=1}^{r^{m-1}} w_{il}\frac{\partial e^m_{i | l}}{\partial o^m_i}.
\end{equation}
Although the network is feed-forward, there are non-linear interactions between S-PSPs. The second term of (\ref{dadw}) captures the hidden dependency of the S-PSPs on the post-synaptic firing count $o^m_i$.

\begin{wrapfigure}{r}{0.3\textwidth}
    \vspace{-10mm}
    \begin{center}
        \includegraphics[width=0.3\textwidth]{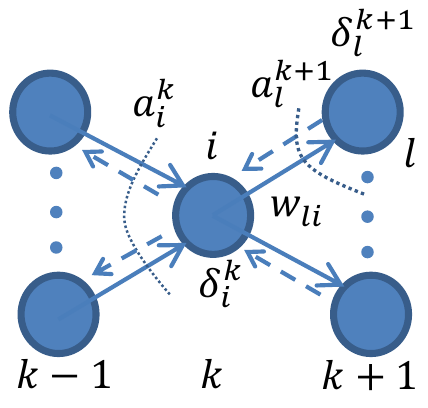}
    \end{center}
    \vspace{-3mm}
    \caption{Macro-level backpropagation at a hidden layer.}
    \label{hidden_bp_img}
\end{wrapfigure}

\textbullet\ \textbf{Hidden layers:} For the $i_{th}$ neuron in the hidden layer $k$, we have
\begin{equation}
    \label{dEdW_hidden}
    \frac{\partial E}{\partial w_{ij}} = \underbrace{\frac{\partial E}{\partial a_i^k}}_\text{macro-level bp} \times \underbrace{\frac{\partial a^k_i}{\partial w_{ij}}}_\text{micro-level bp} = \delta^k_i ~~ \frac{\partial a^k_i}{\partial w_{ij}}.
\end{equation}
The macro-level error backpropagation at a hidden layer is much more involved as in Fig.~\ref{hidden_bp_img}
\begin{equation}
    \label{delta_hidden}
    \delta^k_i = \frac{\partial E}{\partial a^k_i} = \sum_{l=1}^{r^{k+1}}~\frac{\partial E}{\partial a^{k+1}_l} ~\frac{\partial a^{k+1}_l}{\partial a^k_i} = \sum_{l=1}^{r^{k+1}} \delta^{k+1}_l ~\frac{\partial a^{k+1}_l}{\partial a^k_i}.
\end{equation}
According to (\ref{ai_formula}) , (\ref{effect_formula}) and (\ref{tf_formula}), we unwrap $a^{k+1}_l$ and get
\begin{equation}
    a^{k+1}_l = \sum_{p=1}^{r^k} w_{lp}~e^{k+1}_{l | p} =  \sum_{p=1}^{r^k} w_{lp}~f(o^k_p, o^{k+1}_l,  \mathbf{t}_p^{(f)}, \mathbf{t}_l^{(f)}) = \sum_{p=1}^{r^k} w_{lp}~f(g(a^k_p), o^{k+1}_l,  \mathbf{t}_p^{(f)}, \mathbf{t}_l^{(f)}).
\end{equation}
Therefore,  $\frac{\partial a^{k+1}_l}{\partial a^k_i}$ becomes
\begin{equation}
    \label{alai}
    \frac{\partial a^{k+1}_l}{\partial a^k_i} = w_{li}~\frac{\partial e^{k+1}_{l|i}}{\partial o^k_i} ~\frac{\partial o^k_i}{\partial a^k_i} = w_{li}~\frac{\partial e^{k+1}_{l|i}}{\partial o^k_i} g^\prime (a^k_i) = \frac{w_{li}}{\nu}~\frac{\partial e^{k+1}_{l|i}}{\partial o^k_i},
\end{equation}
where the dependency of $e^{k+1}_{l|i}$ on the pre-synaptic firing count $o^k_i$  is considered but the one on the firing timings are ignored, which is supported by the decoupled S-PSP model in (\ref{effect_approx}). Plugging  (\ref{alai}) into (\ref{delta_hidden}), we have
\begin{equation}\label{eqn_final_delta}
    \delta^k_i = \frac{1}{\nu} \sum_{l=1}^{r^{k+1}} \delta^{k+1}_l w_{li} \frac{\partial e^{k+1}_{l|i}}{\partial o^k_i}.
\end{equation}

The micro-stage backpropagation at hidden layers is identical to that at the output layer, i.e. (\ref{dadw}). %The second term of (\ref{dEdW_hidden}) can be obtained similar to (\ref{dadw}). 
Finally, we obtain the derivative of $E$ with respect to $w_{ij}$ as follows
\begin{equation}
    \label{dEdW_mid}
    \frac{\partial E}{\partial w_{ij}} = \delta^k_i e^k_{i | j} \left(1  + \frac{1}{\nu}~\sum_{l=1}^{r^{k-1}} w_{il}\frac{\partial e^k_{i | l}}{\partial o^k_i}\right),
    %\delta^k_i = \begin{cases}
    %\frac{o^m_i - y^m_i}{\nu} &\text{for output layer,}\\
    %\frac{1}{\nu} \sum_{l=1}^{r^{k+1}} \delta^{k+1}_l w_{li} \frac{\partial %e^{k+1}_{l|i}}{\partial o^k_i} &\text{for hidden layers}
    %\end{cases}
\end{equation}
where
\begin{equation}
\delta^k_i = \begin{cases}
\frac{o^m_i - y^m_i}{\nu} &\text{for output layer,}\\
\frac{1}{\nu} \sum_{l=1}^{r^{k+1}} \delta^{k+1}_l w_{li} \frac{\partial e^{k+1}_{l|i}}{\partial o^k_i} &\text{for hidden layers.}
\end{cases}
\end{equation}

Unlike \cite{lee2016training,wu2017spatio}, here decomposing the rate-coded error backpropagation into the macro and micro levels enables computation of the gradient of the actual loss function with respect to the tunable weights, leading to highly competitive performances. Our HM2-BP algorithm can introduce/remove multiple spikes  by one update, greatly improving learning efficiency in comparison with SpikeProp~\cite{bohte2002error}. 
To complete the derivation of HM2-BP, derivatives in the forms of $\frac{\partial e^k_{i|j}}{\partial o^k_i}$ and $\frac{e^k_{i|j}}{\partial o^k_j}$ as needed in (\ref{dadw}) and (\ref{eqn_final_delta}) are yet to be estimated, which is non-trivial as shall be presented in Section~\ref{sec:decouple}.

\subsection{Decoupled Micro-Level Model for S-PSP}
\label{sec:decouple}
The derivatives of the S-PSP $e^k_{i | j}$ with respect to the pre and post-synaptic neuron firing counts are key components in our HM2-BP rule. According to (\ref{effect}), the S-PSP $e^k_{i | j}$ is dependent on both rate and temporal information of the pre and post-synaptic spikes. The firing counts of pre and post-synaptic neurons (i.e., the rate information) are represented by the two nested summations in (\ref{effect}). The exact firing timing information determines the (normalized) post-synaptic potential $\epsilon$ of each pre/post-synaptic spike train pair as seen from (\ref{psp}). The rate and temporal information of spike trains are strongly coupled together, making the exact computation of $\frac{\partial e^k_{i|j}}{\partial o^k_i}$ and $\frac{e^k_{i|j}}{\partial o^k_j}$ challenging. 

To address this difficulty, we propose a decoupled model for $e^k_{i | j}$ to untangle the rate and temporal effects. The model is motivated by the observation that $e^k_{i | j}$ is linear in both $o^k_j$ and $o^k_i$ in the limit of high firing counts. For finite firing rates, we decompose  $e^k_{i | j}$ into an asymptotic rate-dependent effect using the product of $o^k_j$ and $o^k_i$ and a correction factor $\Hat{\alpha}$ accounting for temporal correlations between the pre and post-synaptic spike trains 
\begin{equation}
    \label{effect_approx}
    e^k_{i|j} = \Hat{\alpha}(\mathbf{t}_j^{(f)}, \mathbf{t}_i^{(f)}) o^k_j ~ o^k_i.
\end{equation}
$\Hat{\alpha}$ is a function of exact spike timing. Since the SNN is trained incrementally with small weight updates set by a well-controlled learning rate, $\Hat{\alpha}$ does not change substantially by one training iteration.  Therefore, we approximate  $\Hat{\alpha}$ by using the values of $e^k_{i|j}$, $o^k_j$, and $o^k_i$ available before the next training update by 
\begin{equation*}
    \Hat{\alpha}(\mathbf{t}_j^{(f)}, \mathbf{t}_i^{(f)}) \approx \frac{e^k_{i|j}}{o^k_j o^k_i}.
\end{equation*}
With the micro-level temporal effect considered by $\Hat{\alpha}$,  
%By adopting this decoupled model, we empirically linearize $e^k_{i|j}$ in the total number of spike pairs. This makes sense because the term $e^k_{i|j}$ is mainly dominated by the nested summation, which is in the order of $o^k_j\cdot o^k_i$. If $o^k_j$ and $o^k_i$ are sufficiently large, adding or removing a couple of spikes might not change this averaged temporal effect $\Hat{\alpha}$ too much. We can ensure the large firing counts by setting a reasonably high desired firing level $d_{lvl}$ of the output layer to bring up the firing level of the entire network. And limiting the amount of weight change by setting a small learning rate shall make sure that each time the number of introduced or removed spikes is small. Therefore, under these conditions, we might safely assume that the averaged temporal effect $\Hat{\alpha}$ is a constant within a small range of the given firing counts.
 we estimate the two derivatives by % $\frac{\partial e^k_{i|j}}{\partial o^k_i}$ and $\frac{e^k_{i|j}}{\partial o^k_j}$ by
\begin{align*}
    \frac{\partial e^k_{i|j}}{\partial o^k_i} \approx \Hat{\alpha}~ o^k_j  = ~ \frac{e^k_{i|j}}{o^k_i},~~~~
    \frac{\partial e^k_{i|j}}{\partial o^k_j} \approx \Hat{\alpha}~ o^k_i = ~ \frac{e^k_{i|j}}{o^k_j}.
\end{align*}

%The derivative of the error $E$ with respect to the weight $w_{ij}$ then becomes
%\begin{equation}
%    \label{dEdW_final}
%    \frac{\partial E}{\partial w_{ij}} = \delta^k_i e^k_{j | i} (1  + \frac{1}{\nu}~\sum_{l=1}^{r^{k-1}} w_{il}\frac{e^k_{l | i}}{o^k_i}),
%\end{equation}
%where $\delta^k_i$ follows
%\begin{equation}
%\delta^k_i \approx \begin{cases}
%\frac{o^m_i - y^m_i}{\nu} &\text{for output layer,}\\
%\frac{1}{\nu} \sum_{l=1}^{r^{k+1}} \delta^{k+1}_l w_{li} \frac{e^{k+1}_{i|l}}{o^k_i} &\text{for hidden layers.}
%\end{cases}
%\end{equation}

Our hybrid training method follows the typical backpropagation methodology. First of all, a forward pass is performed by analytically simulating the LIF model (\ref{lif}) layer by layer. Then the firing counts of the output layer are compared with the desirable firing levels to compute the macro-level error. After that, the error in the output layer is propagated backwards at both the macro and micro levels to determine the gradient. Finally, an optimization method (e.g. Adam~\cite{kingma2014adam}) is used to update the network parameters given the computed gradient. 

\section{Experiments and Results}
%Using the proposed error backpropagation rule, we train both deep feed-forward and convolution SNNs for MNIST~\cite{lecun1998gradient} and dynamic N-MNIST~\cite{orchard2015converting} and demonstrate the best performance compared with the state-of-the-art learning algorithms for SNNs~\cite{lee2016training, wu2017spatio}.
\vspace{-3mm}
\paragraph{Experimental Settings and Datasets}
%In this subsection, we briefly describe the experimental settings and datasets used for our experiments. 
The weights of the experimented SNNs are randomly initialized by using the uniform distribution $U[-a, a]$, where $a$ is $1$ for fully connected layers and $0.5$ for convolutional layers. We use fixed firing thresholds  in the range of $5$ to $20$ depending on the layer. We adopt the exponential weight regularization scheme in~\cite{lee2016training} and introduce the lateral inhibition in the output layer to speed up training convergence~\cite{lee2016training}, which slightly modifies the gradient computation for the output layer (see \textbf{Supplementary Material}). We use Adam~\cite{kingma2014adam} as the optimizer and its parameters are set according to the original Adam paper. We  impose greater sample weights for incorrectly recognized data points during the training as a supplement to the Adam optimizer. More training settings are reported in the released source code. % %The learning rate is decayed by $1/\sqrt{i}$ for $i_{th}$ epoch. Other detailed settings can be referred to our code. use the AdaBoost~\cite{freund1997decision} to

The MNIST handwritten digit dataset~\cite{lecun1998gradient} consists of 60k samples for training and 10k for testing, each of which is a $28\times 28$ grayscale image. We convert each pixel value of a MNIST image into a spike train using Poisson sampling based on which the probability of spike generation is proportional to the pixel intensity. The N-MNIST dataset~\cite{orchard2015converting} is a neuromorphic version of the MNIST dataset generated by tilting a Dynamic Version Sensor (DVS)~\cite{lichtsteiner2008} in front of static digit images on a computer monitor. The movement induced pixel intensity changes at each location are encoded as spike trains. Since the intensity can either increase or decrease, two kinds of ON- and OFF-events spike events are recorded. Due to the relative shifts of each image, an image size of $34\times 34$ is produced. Each sample of the N-MNIST is a spatio-temporal pattern with $34 \times 34 \times 2$ spike sequences lasting for $300ms$. We reduce the time resolution of the N-MNIST samples by $600$x to speed up simulation. The Extended MNIST-Balanced (EMNIST)~\cite{cohen2017emnist} dataset, which includes both letters and digits, is more challenging than MNIST. EMNIST has 112,800 training and 18,800 testing samples for 47 classes. We convert and encode EMNIST in the same way as we do for MNIST. We also use the 16-speaker spoken English letters of TI46 Speech corpus~\cite{TI46} to benchmark our algorithm for demonstrating its capability of handling spatio-temporal patterns. There are 4,142  and 6,628 spoken English letters for training and testing, respectively. The continuous temporal speech waveforms are first preprocessed by Lyon's ear model~\cite{lyon1982computational} and then encoded into 78 spike trains using the BSA algorithm~\cite{schrauwen2003bsa}.

We train each network for 200 epochs except for ones used for EMNIST, where we use 50 training epochs. The best recognition rate of each setting is collected and each experiment is run for at least five times to report the error bar. For each setting, we also report the best performance over all the conducted experiments.
\vspace{-3mm}
\paragraph{Fully Connected SNNs for the Static MNIST}
Using Poisson sampling, we encode each $28\times 28$ image of the MNIST dataset into a 2D $784\times L$ binary matrix, where $L=400 ms$ is the duration of each spike sequence, and a $1$ in the matrix represents a spike. The simulation time step is set to be $1ms$. No pre-processing or data augmentation is done in our experiments.
Table~\ref{mnist_table} compares the performance of SNNs trained by the proposed HM2-BP rule with other algorithms. HM2-BP achieves $98.93\%$ test accuracy, outperforming  STBP~\cite{wu2017spatio}, which is the best previously reported  algorithm for fully-connected SNNs. The proposed rule also achieves the best accuracy earlier than  STBP (100 epochs v.s. 200 epochs). We attribute the overall improvement to the hybrid macro-micro processing  that handles the temporal effects and discontinuities at two levels in a way such that explicit back-propagation of the rate-coded error becomes possible and practical. 
%more accurate modeling since the spike discontinuity is implicitly handled within the spike train level.
%We use the entire training set of the MNIST to train and the 10k-sample testing set for reporting the results. 
%Fig.~\ref{output_spikes} shows the output layer spike patterns of the inference on the digit 7 before and after training a $784-800-10$ network. No specific firing patterns can be found initially, but after training the corresponding $7_{th}$ neuron fires most actively while other neurons remain silent, suggesting the correct classification. 
 %Notice that our SNN also performs better than the ANNs trained with Dropout or Drop-connect~\cite{srivastava2014dropout, wan2013regularization}.
\begin{table}[ht]
\centering
\caption{Comparison of different SNN models on MNIST}
\label{mnist_table}
\begin{tabular}{@{}lllll@{}}
\toprule
Model            & Hidden layers  & Accuracy     &Best   & Epochs \\ \midrule
%Standard ANN~\cite{simard2003best}           & 800-800          & 98.4\%          & >100  \\
%Dropout ANN~\cite{srivastava2014dropout}      & 1024-1024-1024  & 98.75\%         & -  \\
%Drop-connect ANN~\cite{wan2013regularization} & 800-800         & 98.88\%         & 1020  \\ %\midrule
Spiking MLP (converted\textsuperscript{*})~\cite{o2013real}      & 500-500    & 94.09\% & 94.09\% & 50         \\ 
Spiking MLP (converted\textsuperscript{*})~\cite{hunsberger2015spiking}       & 500-200    & 98.37\% & 98.37\% & 160               \\
Spiking MLP (converted\textsuperscript{*})~\cite{diehl2015fast}      & 1200-1200               & 98.64\% &98.64\% & 50         \\
Spiking MLP ~\cite{o2016deep}       & 300-300                 & 97.80\% & 97.80\% & 50               \\
Spiking MLP ~\cite{lee2016training}      & 800             & 98.71\%\textsuperscript{a} & 98.71\% &200 \\
Spiking MLP (STBP)~\cite{wu2017spatio}      & 800                     & 98.89\% & 98.89\% & 200           \\
Spiking MLP (this work)      & 800                     & \textbf{98.84} $\pm$ \textbf{0.02\%}   & \textbf{98.93\%} & \textbf{100}\\ \bottomrule
\multicolumn{5}{l}{\footnotesize{We only compare SNNs without any pre-processing (i.e., data augmentation) except for~\cite{o2013real}.}} \\
\multicolumn{5}{l}{\footnotesize{\textsuperscript{*} means the model is converted from an ANN. \textsuperscript{a}~\cite{lee2016training} achieves $98.88\%$ with hidden layers of 300-300.}}
\end{tabular}
\end{table}
\vspace{-4mm}

\paragraph{Fully Connected SNNs for N-MNIST}
The simulation time step is $0.6ms$ for N-MNIST. Table~\ref{nmnist_table} compares the results obtained by different models on N-MNIST. The first two results are obtained by the conventional CNNs with the frame-based method, which accumulates spike events over short time intervals as snapshots and recognizes digits based on sequences of snapshot images. The relative poor performances of the first two models may be attributed to the fact that the frame-based representations tend to be blurry and do not fully exploit spatio-temporal patterns of the input. The two non-spiking LSTM models, which are trained directly on spike inputs, do not perform too well, suggesting that LSTMs may be incapable of dealing with asynchronous and sparse spatio-temporal spikes. The SNN trained by our proposed approach naturally processes spatio-temporal spike patterns, achieving the start-of-the-art accuracy of $98.88\%$, outperforming the previous best ANN ($97.38\%$) and SNN ($98.78\%$) with significantly less training epochs required. 

\begin{table}[ht]
\centering
\caption{Comparison of different models on N-MNIST}
\label{nmnist_table}
\begin{tabular}{@{}lllll@{}}
\toprule
Model            & Hidden layers  & Accuracy    & Best     & Epochs\\ \midrule
Non-spiking CNN~\cite{neil2016phased}      & -    & $95.02\pm 0.30\%$  & -    &-    \\ 
Non-spiking CNN~\cite{neil2016effective}   & -    & $98.30\%$  & 98.30\% & 15-20      \\
Non-spiking LSTM~\cite{neil2016phased}      & -    & $96.93 \pm 0.12 \%$ & - &-         \\
Non-spiking Phased-LSTM~\cite{neil2016phased}  & -    & $97.28 \pm 0.10 \%$ & - &-   \\ \midrule
Spiking CNN (converted\textsuperscript{*})~\cite{neil2016effective}       & -    & 95.72\%  & 95.72\% & 15-20               \\
Spiking MLP ~\cite{cohen2016skimming} & 10000 & 92.87\% & 92.87\% & - \\
Spiking MLP ~\cite{lee2016training}      & 800             & 98.74\% & 98.74\% &200 \\
Spiking MLP (STBP)~\cite{wu2017spatio}      & 800          & 98.78\% & 98.78\% & 200           \\
Spiking MLP (this work)      & 800                     & \textbf{98.84 $\pm$ 0.02\%} & \textbf{98.88\%}  & \textbf{60}\\ \bottomrule
\multicolumn{5}{l}{\footnotesize{Only structures of SNNs are shown for clarity.\textsuperscript{*} means the SNN model is converted from an ANN.}} \\
\end{tabular}
\end{table}
\vspace{-3mm}

\paragraph{Spiking Convolution Network for the Static MNIST}
We construct a spiking CNN consisting of two $5\times5$ convolutional layers with a stride of 1, each followed by a $2\times2$ pooling layer, and one fully connected hidden layer. The neurons in the pooling layer are simply LIF neurons, each of which connects to $2\times2$ neurons in the preceding convolutional layer with a fixed weight of $0.25$. Similar to~\cite{lee2016training,wu2017spatio}, we use elastic distortion~\cite{simard2003best} for data augmentation. As shown in Table~\ref{spikingcnn_table}, our proposed method achieves an accuracy of $99.49\%$, surpassing the best previously reported performance~\cite{wu2017spatio} with the same model complexity after 190 epochs.
\begin{table}[ht]
\centering
\caption{Comparison of different spiking CNNs on MNIST}
\label{spikingcnn_table}
\begin{tabular}{@{}llll@{}}
\toprule
Model            & Network structure        & Accuracy  & Best\\ \midrule
Spiking CNN (converted\textsuperscript{a})~\cite{diehl2015fast}      &  12C5-P2-64C5-P2-10    & 99.12\%    &  99.12\%      \\ 
Spiking CNN (converted\textsuperscript{b})~\cite{esser2015backpropagation}      & -     &92.70\%\textsuperscript{c}    & 92.70\%      \\ 
Spiking CNN (converted\textsuperscript{a})~\cite{rueckauer2017conversion}      & -     &99.44\%   &  99.44\%      \\ \midrule
Spiking CNN ~\cite{lee2016training}      & 20C5-P2-50C5-P2-200-10             & 99.31\% & 99.31\% \\
Spiking CNN (STBP)~\cite{wu2017spatio}      &  15C5-P2-40C5-P2-300-10                     & 99.42\%     &  99.42\%      \\
Spiking CNN (this work\textsuperscript{d})      & 15C5-P2-40C5-P2-300-10                     & 99.32 $\pm$ 0.05\% & 99.36\%  \\ 
Spiking CNN (this work)      & 15C5-P2-40C5-P2-300-10                     & \textbf{99.42} $\pm$ \textbf{0.11\%} & \textbf{99.49\%}  \\ \bottomrule
\multicolumn{4}{l}{\footnotesize{\textsuperscript{a} converted from a trained ANN. \textsuperscript{b} converted from a trained probabilistic model with binary weights. }} \\
\multicolumn{4}{l}{\footnotesize{\textsuperscript{c} performance of a single spiking CNN. 99.42\% obtained for ensemble learning of 64 spiking CNNs.}}\\
\multicolumn{4}{l}{\footnotesize{\textsuperscript{d} performance without data augmentation.}}
\end{tabular}
\vspace{-3mm}
\end{table}
\vspace{-3mm}

\paragraph{Fully Connected SNNs for EMNIST}
Table~\ref{emnist_table} shows that the HM2-BP outperforms the non-spiking ANN and the spike-based backpropagation (eRBP) rule reported in~\cite{neftci2017event} significantly with less training epochs.  
\begin{table}[ht]
\centering
\caption{Comparison of different models on EMNIST}
\label{emnist_table}
\begin{tabular}{@{}lllll@{}}
\toprule
 Model & Hidden Layers   & Accuracy & Best  & Epochs       \\ \midrule
 ANN ~\cite{neftci2017event}& 200-200   & $81.77\%$ & $81.77\%$  & 30       \\  
Spiking MLP (eRBP)~\cite{neftci2017event} & 200-200    & $78.17\%$      & $78.17\%$ & 30         \\ 
Spiking MLP (HM2-BP) & 200-200    & \textbf{84.31} $\pm$ \textbf{0.10\%}&  \textbf{84.43\%} &  \textbf{10}\\
Spiking MLP (HM2-BP)& 800    & \textbf{85.41} $\pm$ \textbf{0.09\%} & \textbf{85.57\%} &   \textbf{19}   \\ \bottomrule
\end{tabular}
\end{table}
\vspace{-3mm}

\paragraph{Fully Connected SNNs for TI46 Speech} 
The HM2-BP produces excellent results on the 16-speaker spoken English letters of TI46 Speech corpus~\cite{TI46} as shown in Table~\ref{ti46_table}. This is a challenging spatio-temporal speech recognition benchmark and no prior success based on SNNs was reported.
% Additional benchmark table goes to here:
\begin{table}[ht]
\centering
\caption{Performances of HM2-BP on TI46 (16-speaker speech)}
\label{ti46_table}
\begin{tabular}{@{}llll@{}}
\toprule
Hidden Layers   & Accuracy & Best    &Epochs      \\ \midrule
800   & $89.36\pm 0.30\%$ &$89.92\%$   & 138      \\  
400-400   & $89.83\pm 0.71\%$ & $90.60\%$    & 163     \\  
800-800   & $90.50\pm 0.45\%$ & $90.98\%$   & 174       \\  \bottomrule
\end{tabular}
\end{table}
\vspace{-4mm}

\begin{figure*}[ht]
    \vspace{-3mm}
    \begin{center}
        \includegraphics[width=0.9\textwidth]{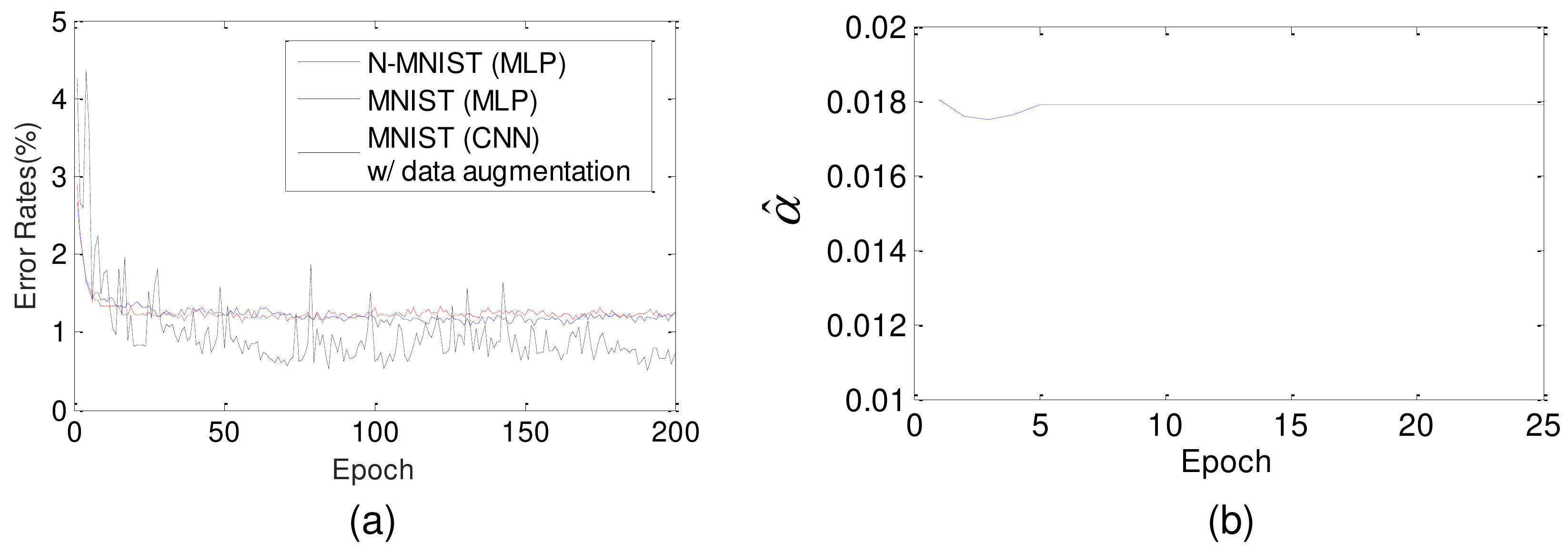}
    \end{center}
    \vspace{-3mm}
    \caption{(a) HM2-BP convergence for the first three reported experiments; (b) $\hat{\alpha}$ v.s. epoch.}
    \label{covergence}
    \vspace{-3mm}
\end{figure*}

\paragraph{In-depth Analysis of the MNIST and N-MNIST Results}
Fig.~\ref{covergence}(a) plots the HM2-BP convergence curves for the best settings of the first three experiments reported in the paper. The convergence is logged in the code. Data augmentation contributes to the fluctuation of convergence in the case of Spiking Convolution network. We conduct the experiment to see if our assumption used in approximating $\hat{\alpha}$ of (\ref{effect_approx}) is valid. Fig.~\ref{covergence}(b) shows that the value of $\hat{\alpha}$ of a randomly selected synapse does not change substantially over epochs during the training of a two-layer SNN (10 inputs and 1 output). At the high firing frequency limit, the S-PSP is proportional to $o_j^k \cdot o_i^k$, making the multiplicative dependency on the two firing rates a good choice in (25).  

%We want to emphasize on the strong capability of HM2-BP for handling the spatio-temporal inputs. N-MNIST is already a spatio-temporal benchmark for which the HM2-BP outperforms the best reported result even considering its error bar. Table \ref{performance_table} includes two additional datasets. Extended MNIST (EMNIST)~\cite{cohen2017emnist} is more challenging than MNIST for which the HM2-BP outperforms the non-spiking ANN and the spike-based backpropagation (ERBP) reported in \cite{neftci2017event} significantly. The HM2-BP produces excellent results on the 16-speaker spoken English letter of TI46 Speech corpus~\cite{TI46}. This is a challenging spatio-temporal speech recognition benchmark and no prior success based on SNNs was reported before. 

\paragraph{Training Complexity Comparison and Implementation}
Unlike~\cite{wu2017spatio}, our hybrid method does not unwrap the gradient computation in the time domain, roughly making it $O(N_T)$ times more efficient than~\cite{wu2017spatio}, where $N_T$ is the number of time points in each input example. %Our hybrid rule performs one gradient246computation per weight for each sample, unlike [27] that unwraps the gradient computation in the247temporal domain, whose complexity roughly in the order ofO(NT), whereNTis the number of time248points in each sample
%Table~\ref{training_complexity} compares the complexity of the cutting-edge direct SNN training algorithms~\cite{lee2016training,wu2017spatio} with the proposed approach. The complexity here is defined as the number of gradients with respect to weights is required for the update of each synaptic weight. The previous state-of-the-art~\cite{wu2017spatio} needs to compute the gradient at each time step (see Equ. (22) and (23) in~\cite{wu2017spatio}), leading to a significant overhead compared with our rule. Although~\cite{lee2016training} is in the same complexity level as the proposed method, our method outperforms theirs because of less approximation used and better management of discrete spike events. 
The proposed method can be easily implemented. We have made our CUDA implementation available online\footnote{\url{https://github.com/jinyyy666/mm-bp-snn}}, the first publicly available high-speed GPU framework for direct training of deep SNNs. %It is worthwhile mentioning that no one has ever published the code before due to the high complexity of training SNN models.
%, and achieves a good throughput of 100 images/sec on a NVIDIA Titan XP GPU. It is worthwhile mentioning that no one has ever reported the training throughput before due to the high complexity of training SNN models. 

\section{Conclusion and Discussions}
In this paper, we present a novel hybrid macro/micro level error backpropagation scheme to train deep SNNs directly based on spiking activities. The spiking timings are exactly captured in the spike-train level post-synaptic potentials (S-PSP) at the microscopic level. The rate-coded error is defined and efficiently computed and back-propagated across both the macroscopic and microscopic levels. We further propose a decoupled S-PSP model to assist gradient computation at the micro-level. In contrast to the previous methods, our hybrid approach directly computes the gradient of the rate-coded loss function with respect to tunable parameters. Using our efficient GPU implementation of the proposed method, we demonstrate the best performances for both fully connected and convolutional SNNs over the static MNIST, the dynamic N-MNIST and the more challenging EMNIST and 16-speaker spoken English letters of TI46 datasets, outperforming the best previously reported SNN training techniques. Furthermore, the proposed approach also achieves competitive performances better than those of the conventional deep learning models when dealing with asynchronous spiking streams.  

The performances achieved by the proposed BP method may be attributed to the fact that it addresses key challenges of SNN training in terms of scalability, handling of temporal effects, and gradient computation of loss functions with inherent discontinuities. Coping with these difficulties through error backpropagation  at both the macro and micro levels provides a unique perspective to  training of SNNs. More specifically, orchestrating the information flow based on a combination of temporal effects and firing rate behaviors  across the two levels in an interactive manner allows for the definition of the rate-coded loss function at the macro level, and backpropagation of errors from the macro level to the micro level, and back to the macro level.  This paradigm provides a practical solution to the difficulties brought by discontinuities inherent in an SNN while capturing  the micro-level timing information via S-PSP.  As such, both rate and temporal information in the SNN is  exploited during the training process, leading to the state-of-the-art performances. By releasing the GPU implementation code in the future, we expect this work would help move the community forward towards enabling  high-performance spiking neural networks and neuromorphic computing. 

%Unlike the previous methods that explicitly tackle discontinuity of each spike, our method bypasses this necessity by adopting the S-PSP, which is a continuous variable defined in the entire spike train level. By theoretical derivation, we show that a direct gradient of the objective with respect to the weight can be obtained under the proposed rule. We propose an average model of S-PSP to decouple temporal and rate effect for feasibly getting more accurate gradients. 

\subsubsection*{Acknowledgments}
This material is based upon work supported by the National Science Foundation under Grant No.CCF-1639995 and the Semiconductor Research Corporation (SRC) under Task 2692.001. The authors would like to thank High Performance Research Computing (HPRC) at Texas A\&M University for providing computing support. Any opinions, findings, conclusions or recommendations expressed in this material are those of the authors and do not necessarily reflect the views of NSF, SRC, Texas A\&M University, and their contractors.

\small
\bibliographystyle{plain}
\bibliography{neurips_2018}

\newpage
\appendix
\renewcommand{\theequation}{A-\arabic{equation}}
\renewcommand{\thefigure}{A-\arabic{figure}}
\section*{Supplementary Material}
\section{Gradient Computation for the Output Layer with Lateral Inhibition}
\begin{wrapfigure}{r}{0.35\textwidth}
    \begin{center}
        \includegraphics[width=0.2\textwidth]{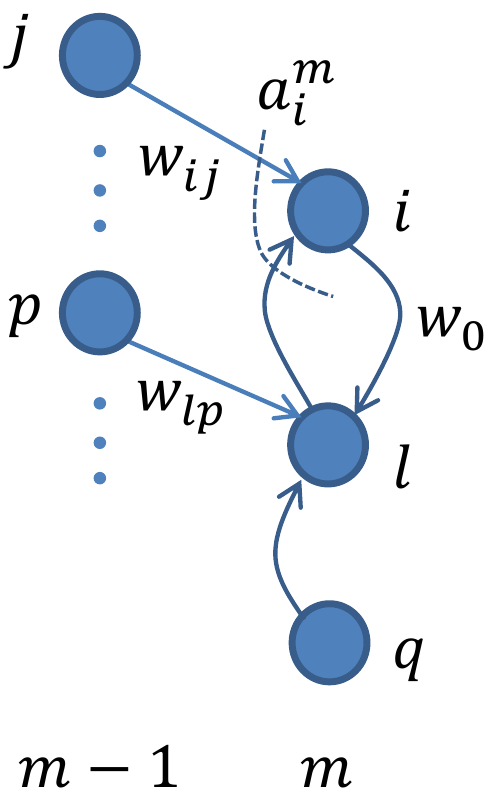}
    \end{center}
    \caption{The output layer with lateral inhibition.}
    \label{lateral_img}
    \vspace{-3mm}
\end{wrapfigure}

Without loss of generality, it is assumed that lateral inhibition exists between every pair of two output neurons. It is also assumed that the weights for lateral inhibition are all fixed at the constant value of $w_0$. As shown in Fig.~\ref{lateral_img}, the total post-synaptic potential (T-PSP) $a^m_i$ for the neuron $i$ at the output layer $m$ can be unwrapped as
\begin{equation*}
    a^m_i = \sum_{j = 1}^{r^{m-1}} w_{ij} ~ e^m_{i | j} + \sum_{l\neq i}^{r^m} w_0~ e^m_{i | l},
\end{equation*}
where the second term describing the lateral inhibition effects between neurons in the same output layer.

The derivative of $a^m_i$ with respect to $w_{ij}$ is
\begin{equation}
    \begin{split}\label{lateral_ai}
        \frac{\partial a^m_i}{\partial w_{ij}} & = \frac{\partial }{\partial w_{ij}}\left(\sum_{j = 1}^{r^{m-1}} w_{ij} ~ e^m_{i | j}\right) + \frac{\partial }{\partial w_{ij}}\left(\sum_{l\neq i}^{r^m} w_0~ e^m_{i | l}\right) \\
        & = e^m_{i|j} \left(1 + \frac{1}{\nu}\sum_{h=1}^{r^{m-1}}w_{ih}\frac{\partial e^m_{i|h}}{\partial o^m_i}\right) + \sum_{l\neq i}^{r^m} w_0 \frac{\partial e^m_{i|l}}{\partial o^m_l} g^\prime(a^m_l)\frac{\partial a^m_l}{\partial w_{ij}}.
    \end{split}
\end{equation}
Similarly for the neuron $l$ of the output layer $m$, $a^m_l$ is given as
\begin{equation*}
    a^m_l = \sum_{p = 1}^{r^{m-1}} w_{lp} ~ e^m_{l | p} + \sum_{q\neq l}^{r^m} w_0~ e^m_{l | q}.
\end{equation*}
Therefore, $\frac{\partial a^m_l}{\partial w_{ij}}$ is
\begin{equation}
    \label{lateral_al}
    \frac{\partial a^m_l}{\partial w_{ij}} = w_0 \frac{\partial e^m_{l|i}}{\partial o^m_i} g^\prime(a^m_i) \frac{\partial a^m_i}{\partial w_{ij}}.
\end{equation}
Plugging (\ref{lateral_al}) back into (\ref{lateral_ai}) and solving for $\frac{\partial a^m_i}{\partial w_{ij}}$ leads to
\begin{equation}
    \label{lateral_dadw}
    \frac{\partial a^m_i}{\partial w_{ij}} = \gamma~~e^m_{i|j} \left(1 + \frac{1}{\nu}\sum_{h=1}^{r^{m-1}}w_{ih}\frac{\partial e^m_{i|h}}{\partial o^m_i}\right),
\end{equation}
where
\begin{equation*}
    \gamma = \frac{1}{1 - \frac{w_0^2}{\nu^2}\sum_{l\neq i}^{r^m}\frac{\partial e^m_{i|l}}{\partial o^m_l}\frac{\partial e^m_{l|i}}{\partial o^m_i}}.
\end{equation*}
(\ref{lateral_dadw}) is identical to (\ref{dadw}) except that the factor $\gamma$ is introduced to capture the effect of lateral inhibition.

For the output layer with lateral inhibition, the term $\frac{\partial E}{\partial a^m_i}$ of the macro-level backpropagation defined in (\ref{dEdW}) is the same as the one without lateral inhibition. 
\end{document}